\documentclass[conference]{IEEEtran}

\pdfoutput=1

\IEEEoverridecommandlockouts
\usepackage{cite}
\usepackage{amsmath,amssymb,amsfonts}
\usepackage{algorithmic}
\usepackage{graphicx}
\usepackage{textcomp}
\usepackage{xcolor}
\def\BibTeX{{\rm B\kern-.05em{\sc i\kern-.025em b}\kern-.08em
    T\kern-.1667em\lower.7ex\hbox{E}\kern-.125emX}}

\usepackage{multirow}
\usepackage{hyperref}
\hypersetup{
    colorlinks,
    linkcolor={red!50!black},
    citecolor={blue!50!black},
    urlcolor={blue!80!black}
}

\begin{document}

\title{\textbf{E}fficient \textbf{Ne}ural \textbf{L}ight \textbf{F}ields (ENeLF) for Mobile Devices\\
}


\author{\IEEEauthorblockN{Austin Peng}
\IEEEauthorblockA{\textit{College of Computing} \\
\textit{Georgia Institute of Technology}\\
Atlanta, GA, USA \\
apeng39@gatech.edu}
}

\maketitle

\begin{abstract}
Novel view synthesis (NVS) is a challenge in computer vision and graphics, focusing on generating realistic images of a scene from unobserved camera poses, given a limited set of authentic input images. Neural radiance fields (NeRF) achieved impressive results in rendering quality by utilizing volumetric rendering. However, NeRF and its variants are unsuitable for mobile devices due to the high computational cost of volumetric rendering. Emerging research in neural light fields (NeLF) eliminates the need for volumetric rendering by directly learning a mapping from ray representation to pixel color. NeLF has demonstrated its capability to achieve results similar to NeRF but requires a more extensive, computationally intensive network that is not mobile-friendly. Unlike existing works, this research builds upon the novel network architecture introduced by MobileR2L and aggressively applies a compression technique (channel-wise structure pruning) to produce a model that runs efficiently on mobile devices with lower latency and smaller sizes, with a slight decrease in performance.
\end{abstract}


\section{Introduction}
Impressive breakthroughs in NVS and 3D reconstruction have recently been made. The trained models can generate 3D views from 2D examples, making them highly desirable and versatile (e.g. virtual telepresence, metaverse). However, widespread adoption of such applications requires them to run efficiently on resource-constrained devices. Considering the limitations in resource-constrained devices regarding computing power and storage, there has been an increased interest in research on rendering efficiency. Despite the remarkable rendering quality in the family of NeRF techniques, their slow, repetitive sampling method results in non-real-time performance. Although various approaches have attempted to address this issue, many still rely on high-end GPUs, making them impractical for resource-constrained devices. Emerging research methods rely solely on neural networks that can directly encapsulate the 3D representation from multiple 2D observations.

ENeLF proposes a solution tailored for real-time NVS on mobile devices. Drawing inspiration from previous work like R2L and MobileR2L, ENeLF employs data distillation, an efficient CNN backbone and super-resolution module, and the pruning model compression technique. This design choice enables further real-time rendering at a small sacrifice in quality.

\section{Related Works}
\textbf{NeRF.} NeRF achieved impressive results in rendering quality for view synthesis by mapping a 5D input $(x, y, z, \theta, \phi)$ to 4D output $(R, G, B, \sigma)$ \cite{nerf2021}. However, a notable drawback of NeRF is that the multi-layer perceptron (MLP) is sampling hundreds of times per pixel to perform accurate volumetric rendering, resulting in slow training and inference times for high-resolution images. Researchers have conducted many additional studies utilizing more efficient radiance field representations \cite{nsvf2020, kilonerf2021, snerg2021, pleno2021}, or distilling meshes from the radiance field \cite{mobilenerf2023, bakedsdf2023, r2l2022, mobiler2l2023}. However, few studies have achieved real-time rendering on mobile devices.

\textbf{NeLF.} NeLF is an emerging research field and can achieve similar rendering quality to NeRF with significant inference speedup. NeLF utilizes a light field (integral of radiance) and can reduce the hundreds of samples per ray to a single sample per ray, resolving the bottleneck sampling issue by eliminating volumetric rendering. However, the light field space is more complex than the radiance field. NeLF requires a larger model and more data to achieve a similar rendering quality. The increased computational cost causes higher training time than NeRF but at significantly lower inference time.

\textbf{Real-Time NeLF Rendering.} In achieving real-time rendering speed of NeLF, three main groups of approaches tackle this issue. \textbf{(1)} The first group includes approaches like NeuLF \cite{neulf2021} and LFN \cite{lfn2021}, which frame NVS as directly mapping the camera rays to the pixel color, making inference speed faster by requiring a single forward pass per pixel. However, the light field is more complex than the radiance field, making the learning process more challenging. The intuition for this is that radiance at neighboring spaces does not change significantly (given the radiance field in the physical world is typically continuous). In contrast, two neighboring rays can drastically differ due to occlusion. A second approach of \textbf{(2)} altering network design is taken to resolve this issue with light field complexity. R2L \cite{r2l2022} modified their MLP to a depth of 88 layers (compared to NeRF's 11-layer MLP) to introduce sufficient network representation space to learn the light field. Due to the significant increase in model size, R2L employs a pre-trained teacher NeRF model to distill data and help train their deeper MLP. With their hardware and software setup, R2L achieved $\sim$30$\times$ speedup and $\sim$2 DB PSNR increase compared to NeRF \cite{nerf2021}. \textbf{(3)} The third group found ways to represent the light field space in alternate ways. NeLF with Ray-Space Embeddings (RSE) \cite{rse2022} maps 4D ray-space into an intermediate, interpolable latent space. SIGNET \cite{signet2021} utilizes a variety of Fourier-inspired input transformation strategies to represent the light field. SIGNET's main limitation is that it doesn't guarantee a photorealistic reconstruction \cite{ls2024}. As a result, this research will mainly focus on utilizing the findings of MobileR2L \cite{mobiler2l2023}, which introduces a super-resolution (SR) module into the MLP so the network can learn the pixel colors of missing rays through upsampling. Focus is placed on the techniques utilized by MobileR2L as their SR module introduces significant efficiencies by only needing to pass a fraction of the rays through the model and learning the representation of the remaining pixels via SR. Throughout the paper, comparisons will be made to MobileR2L \cite{mobiler2l2023}, the current state-of-the-art (SOTA) method for real-time rendering on mobile devices, achieving the highest PSNR among all existing techniques.

\section{Methodology}
\subsection{Prerequisites: R2L, MobileR2L, Pruning}
\begin{figure}[htbp]
    \centerline{\includegraphics[width=\linewidth]{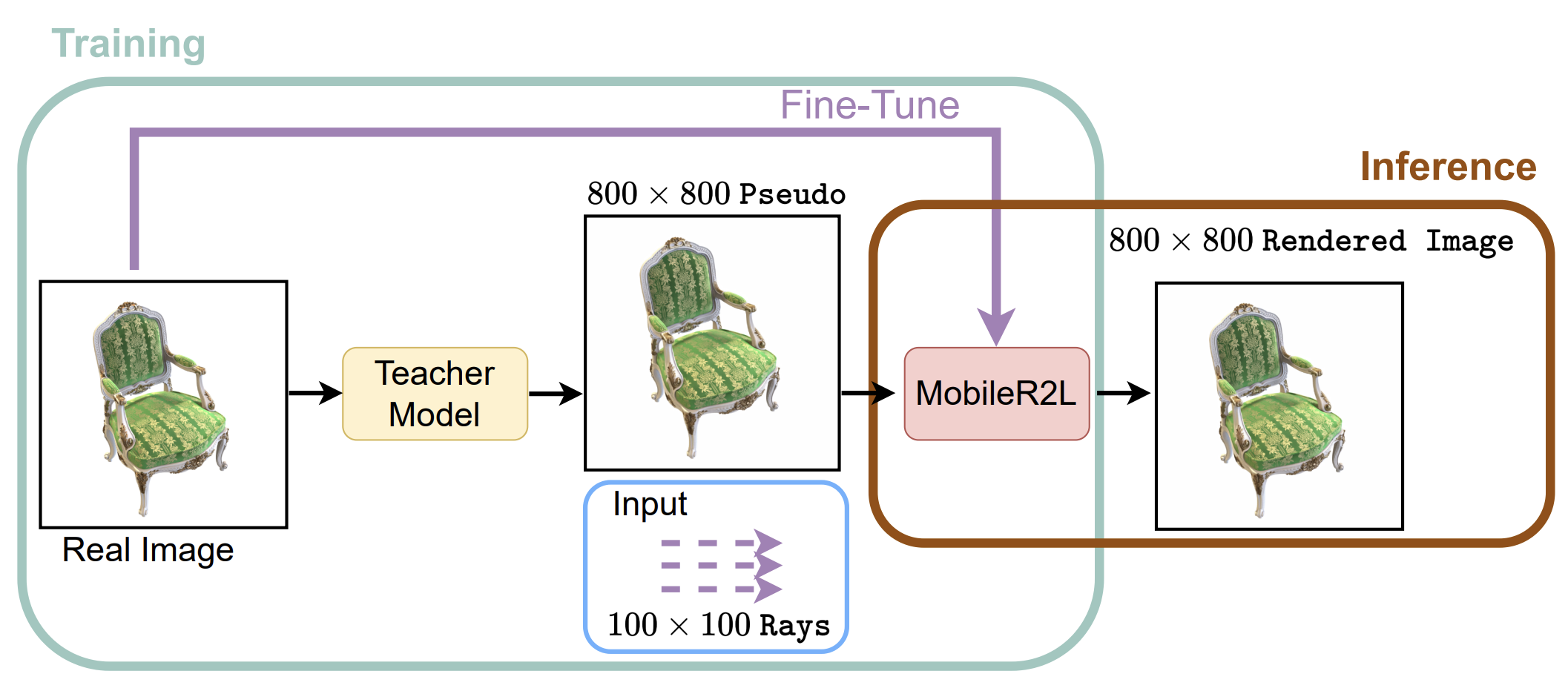}}
    \caption{R2L/MobileR2L training and inference network design.}
    \label{fig:r2l-network-design}
\end{figure}

\textbf{R2L.} NeLF maps the camera ray directly to a pixel color, but because the light field is much harder to learn than the radiance field, R2L \cite{r2l2022} proposes an 88-layer deep residual MLP architecture to learn the mapping from ray to pixel color. Using residuals allows the model architecture to be much deeper and capture the complex light field space. However, a larger model needs more data to learn the representation sufficiently. R2L tackles this issue by pre-training a NeRF model as a teacher model to synthesize additional pseudo data. Then, they adopt a two-stage training process. First, the distilled data is used to train the R2L model. Then, the original data is used to finetune the R2L model. Their technique allowed the R2L model to achieve performance comparable to the teacher NeRF model. Fig. \ref{fig:r2l-network-design} illustrates the training and inference pipeline.

\textbf{MobileR2L.} R2L network design introduced the possibility of real-time rendering and efficiency gains, but it still had high memory costs. For example, an 800$\times$800 image still needs the prediction of 640,000 rays, which causes out-of-memory issues even with high-end GPUs (e.g. A100) \cite{mobiler2l2023}. MobileR2L introduced an SR module that allows the forward pass of a fraction of the rays and learns the rest of the pixels through upsampling with transposed convolutions. In addition, MobileR2L converts the R2L MLP model into a CNN with residual blocks, making the model backbone more efficient and better performing. These two discoveries allowed MobileR2L to set SOTA on the realistic synthetic 360$^\circ$ and real-world forward-facing datasets.

\textbf{Pruning.}
Neural network pruning is widely used to reduce the high inference cost of deep neural networks in resource-constrained settings. Generally, the pruning process consists of three stages: (1) training, (2) pruning, and (3) finetuning. Standard pruning criteria include magnitude-based and scaling factor-based methods. During pruning, extra and unnecessary weights are removed following the selected criterion. Removing redundant weights and retaining high-importance weights allow for a reduction in model size while preserving performance. Though the pruned model can suffer from worse performance, this is the typical tradeoff encountered when dealing with efficiency and performance \cite{pruning2018}. 

Deep convolutional neural networks (CNNs) generally have a high computational cost, causing deployment to be an issue for real-world applications. Network slimming is an approach for (1) reducing CNN model size, (2) decreasing run-time memory, and (3) lowering the compute operations, all with minimal effect on accuracy \cite{networkslimming2017}.

\subsection{ENeLF}

\begin{figure*}[htbp]
    \centerline{\includegraphics[width=1\linewidth]{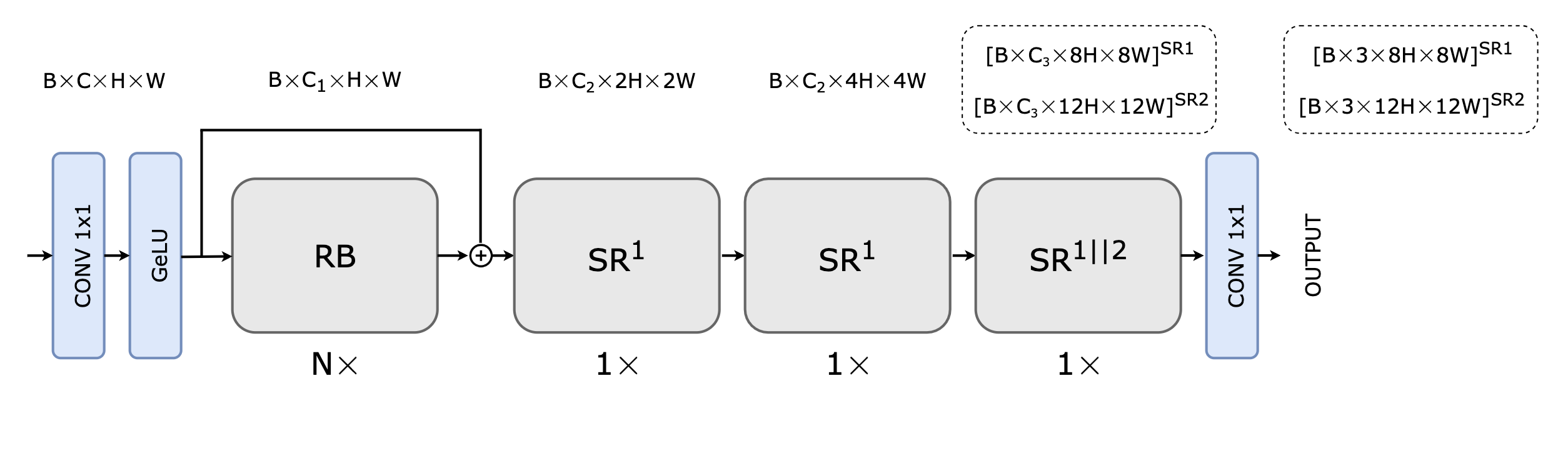}}
    \centerline{\includegraphics[width=1\linewidth]{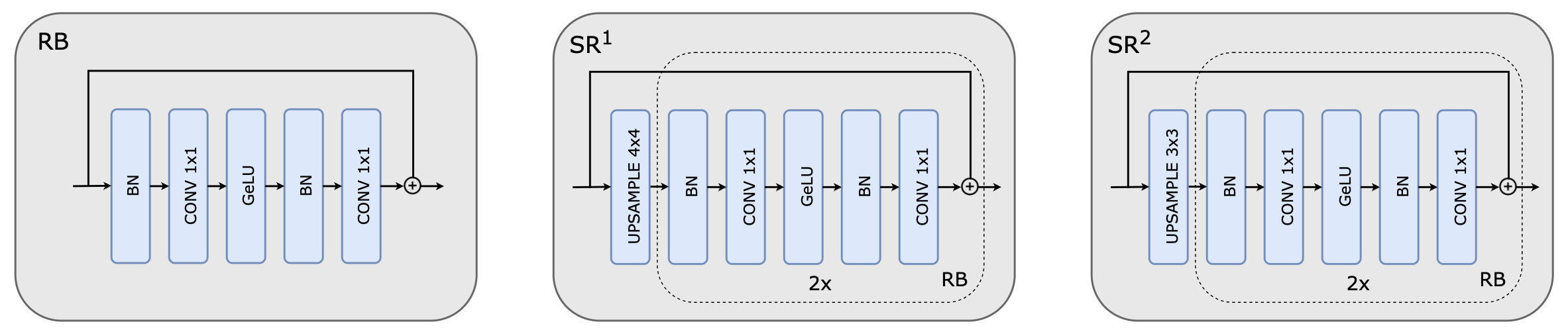}}
    \caption{Modified MobileR2L model architecture to support channel-wise structure pruning.}
    \label{fig:enelf-arch}
\end{figure*}

\textbf{Overview.} The experiments conducted follow the training process of R2L \cite{r2l2022}, using (1) a pre-trained teacher NeRF model \cite{nerf2021} and (2) performing data distillation with that teacher model to generate additional data. The efficient CNN backbone and SR modules in MobileR2L \cite{mobiler2l2023} are leveraged by ENeLF, significantly reducing the model size and inference speed compared to R2L. Modifications to apply efficiency discoveries from these two papers with channel-wise structure pruning are made to the MobileR2L model architecture, transforming it to the ENeLF architectecture.

\textbf{Channel-wise Structure Pruning}
The specific pruning method investigated for CNNs is channel-wise structure pruning, where the convolutional (CONV) layer channels are pruned based on their importance. This approach defines "importance" with the learnable scaling factors parameters in the batch normalization (BN) layers. An L1-norm can be applied to the learnable parameters of BN layers and push them towards 0, enabling the usage of BN scaling factor magnitude to identify the importance of channels. Channels with higher associated scaling factors are more critical and are retained. Channels with smaller associated scaling factors are less important and are pruned. This proposed method requires no special software or hardware accelerator to achieve speedups.

\textbf{Reorder BN and CONV layers.}
The previous section mentioned that BN layer scaling factors are directly responsible for determining which channels of the CONV layer to prune. However, MobileR2L placed their BN layers after the CONV layers. In the implementation of ENeLF, the BN layers are placed before the CONV layers. Experimental results on ENeLF showed that the reordering of these layers did not significantly affect model performance.

\section{Experiments}
\textbf{Datasets.} Most studies perform experiments on two datasets: realistic synthetic 360$^{\circ}$ \cite{nerf2021} and real-world forward-facing \cite{llff2019}. The realistic synthetic 360$^{\circ}$ dataset contains 8-path traced scenes, including 100 images for training and 200 for testing for each scene. Due to GPU constraints, experiments are performed on select scenes (e.g. chair, drums, ficus, lego) from the realistic synthetic 360$^{\circ}$ dataset. Following research conducted in MobileR2L \cite{mobiler2l2023}, the experiments on ENeLF for realistic synthetic 360$^{\circ}$ will also be conducted on the original resolution of the dataset (800 $\times$ 800).

\textbf{Implementation.} Leveraging the best-performing architecture from MobileR2L, D60-SR3 (60 layers, 3 SR blocks), the experimental procedure is as follows: (1) train the teacher NeRF model on select scenes from realistic synthetic 360$^\circ$, (2) distill 5K examples per scene (50\% less compared to MobileR2L), (3) modify MobileR2L to a channel-wise pruning friendly architecture ENeLF, (4) train ENeLF for 200K iterations (33\% less compared to MobileR2L) with batch size of 10, (5) perform channel-wise structure pruning, and (6) retrain ENeLF for 60K iterations. Significant dataset and training time reductions are made due to academic GPU and storage space constraints.

The ENeLF model accepts an input of size 100$\times$100 for the realistic synthetic 360$^\circ$ dataset and upsamples it by
8$\times$ to render an 800$\times$800 image. For the real-world forward-facing dataset, an input of size 84$\times$63 is upsampled 12$\times$ to render a 1008$\times$756 image. Note that CONV layer kernel size and padding are adjusted in the final SR block to achieve the desired 8$\times$ and 12$\times$ upsampling. Realistic synthetic 360$^\circ$ uses 3 SR$^1$ to achieve 8$\times$ upsampling while real-world forward-facing uses 2 SR$^1$ and 1 SR$^2$ to achieve 12$\times$ upsampling.

\begin{figure*}[htbp]
    \centerline{\includegraphics[width=1\linewidth]{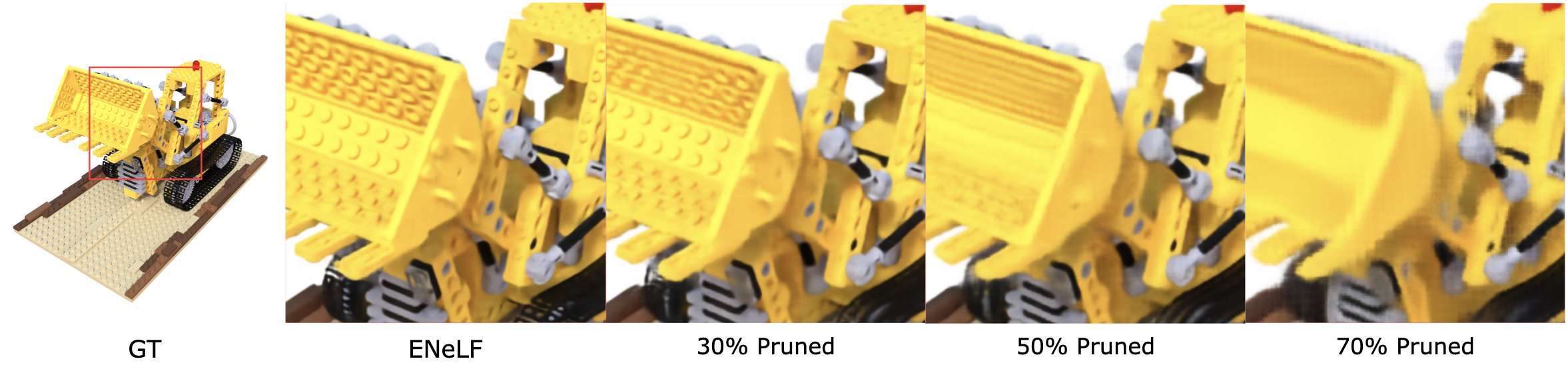}}
    \caption{Qualitative visual comparison between ENeLF and ground truth on the realistic synthetic 360$^\circ$ lego scene of size 800$\times$800. It is best viewed in color.}
    \label{fig:enelf-qualitative-results}
\end{figure*}

\textbf{Model Performance.} Due to the many adaptations made to the experimental procedure in MobileR2L for ENeLF, an attempt is made to reproduce the work of MobileR2L with reduced scenes, 50\% distilled data, 33\% less training epochs, and smaller batch size. The reasoning is to ensure fair and comparable results despite different environments. To understand the rendering quality of MobileR2L and ENeLF, three standard metrics are reported (PSNR \cite{psnr2008}, SSIM \cite{ssim2004}, and LPIPS \cite{lpips2018}) on realistic synthetic 360$^\circ$ dataset.

\begin{table}[htbp]
    \caption{Realistic Synthetic 360$^\circ$}
    \begin{center}
        \begin{tabular}{c|cccc|ccc}
        Pruning Ratio                        & PSNR $\uparrow$ & SSIM $\uparrow$ & LPIPS $\downarrow$ \\ \hline
        0\% - MobileR2L \cite{mobiler2l2023} & \textbf{31.34}         & \textbf{0.993}         & 0.051            \\
        0\% - Mobile R2L Reproduced          & 29.54         & \textbf{0.993}         & \textbf{0.048}            \\ \hline
        30\%                                 & 27.79         & 0.990         & 0.0738           \\
        50\%                                 & 26.71         & 0.989         & 0.094            \\
        70\%                                 & 24.81         & 0.984         & 0.137 
        \end{tabular}
        \label{tab1}
    \end{center}
\end{table}

\begin{table}[htbp]
    \caption{Realistic Synthetic 360$^\circ$}
    \begin{center}
        \begin{tabular}{c|cccc}
        Pruning Ratio                        & \# Params $\downarrow$ & FLOPs $\downarrow$ & Size $\downarrow$ & Latency $\downarrow$ \\ \hline
        0\% - MobileR2L \cite{mobiler2l2023} & 4.3 M                & 148.52 G         & 8.3 MB          & \textbf{22.65 ms}           \\
        0\% - Reproduced                     & 4.3 M                & 148.52 G         & 8.8 MB          & 175.06 ms          \\ \hline
        30\%                                 & 3.11 M               & 124.72 G         & 6.3 MB          & 142.62 ms          \\
        50\%                                 & 2.29 M               & 108.52 G         & 4.8 MB          & 134.39 ms          \\
        70\%                                 & \textbf{1.55 M}               & \textbf{93.8   G}         & \textbf{3.3 MB}          & 116.37 ms
        \end{tabular}
        \label{tab2}
    \end{center}
\end{table}

Model parameters, FLOPs, size (MB), and latency are often used when evaluating efficiency. Although ENeLF achieved worse results on PSNR, SSIM, and LPIPS (see Table \ref{tab1}), it achieved better results on all efficiency metrics (see Table \ref{tab2}). Latency measurements are profiled with an iPhone 15 (iOS 17) using CoreMLTools \cite{coremltools}.

As for qualitative results (see Fig. \ref{fig:enelf-qualitative-results}), the pruned models all lose some capability to render fine-grained details in the resulting views. This is likely due to the reduced parameters in the ENeLF model, which reduces the model's capability to represent the light field space accurately.

\section{Limitations}
The main limitation of this work is training time. Following the procedure from R2L \cite{r2l2022}, 10k additional images were distilled by the teacher NeRF model to train ENeLF. The additional training examples were increased $\sim$10$\times$ compared to the original dataset (up to 100 examples), resulting in longer training times than the NeRF approaches. There are existing approaches that apply pruning at different stages of training to identify winning tickets (small but critical subnetworks) for dense, randomly initialized deep networks \cite{lth2018}. However, identifying winning tickets still requires a costly train, prune, and retrain process. A newer work, Early-Bird Ticket (EBT) \cite{ebt2019}, offers a more efficient approach to finding "winning tickets" by determining an early stopping point for initial training with pruned mask distance. The authors of EBT also found that EBTs consistently emerge in ResNet-like models at early epoch ranges, even when training at a lower precision.

Another limitation is using a widely shared high-traffic, high-performance computing cluster at Georgia Institute of Technology. GPU availability is scarce, and training time for each ENeLF model is high (i.e., 16 GPU hours on H100 for only 200k iterations). Consequently, the results are not as flushed out as they could be. However, these results show significant improvements in model efficiency with minor performance degradations, which could be improved with additional training.

Finally, ENeLF fails to generate high-frequency details in the image, a similar issue to MobileR2L. This is likely due to the reduced representation a smaller pruned model can capture. A larger model may reduce this problem but will impact latency and deployment to mobile devices.

Future research should focus on further compression of ENeLF (e.g. quantization-aware training with static quantization or cyclic precision training \cite{cpt2021}) and improving the pruning process (train, prune, retrain) to help boost the performance and efficiency of the current model.

\section{Conclusion}

This work presents ENeLF, the first instance of applying compression techniques to a NeLF network that renders images with slight performance degradations compared to MobileR2L \cite{mobiler2l2023} and NeRF \cite{nerf2021} while running on model devices. Experiments are performed to prune and compress an optimized network architecture trained using data distillation to render high-resolution images.

\vspace{12pt}
\end{document}